\documentclass{article}
\usepackage{conference,times}
\usepackage{amsmath,amssymb,booktabs,multirow,microtype,graphicx}
\usepackage{hyperref}
\usepackage{url}
\usepackage{algorithm}
\usepackage{algpseudocode}
\usepackage{wrapfig}
\usepackage{tabularx}
\usepackage{colortbl}
\usepackage{array}

\usepackage[table]{xcolor}
\usepackage[most]{tcolorbox}
\definecolor{promptgreen}{HTML}{8BC78A}
\definecolor{promptdarkgreen}{HTML}{5DA05C}
\definecolor{promptbackground}{HTML}{F8FCF7}
\newcommand{\method}{\textsc{TTRSD}}

\title{\method: Test-Time Reinforcement Learning with Self-Distillation for Vision-Language Models}

\author{
\makebox[\textwidth]{\hss\textbf{Shuning Wang$^{1*}$, Zhiheng Wu$^{2*\ddagger}$, Xun Zhou$^{1*}$, Chongyang Cui$^{2}$, Chen Jia$^{2}$, Bowen Liu$^{3}$,}\hss} \\
\makebox[\textwidth]{\hss\textbf{Chuanjie Li$^{4}$, Xiang Chen$^{1}$, Yi Yang$^{2}$, Yumeng Zhang$^{2\dagger}$, Wenjie Huang$^{5\dagger}$}\hss} \\[4pt]
\makebox[\textwidth]{\hss$^{1}$Zhejiang University \quad $^{2}$Baidu.Inc \quad $^{3}$Hong Kong University of Science and Technology\hss} \\
\makebox[\textwidth]{\hss$^{4}$Harbin Institute of Technology \quad $^{5}$Anhui University\hss} \\[2pt]
\makebox[\textwidth]{\hss$^{*}$Equal contribution \quad $^{\dagger}$Corresponding author \quad $^{\ddagger}$Project leader\hss}
}

\finalcopy
\begin{document}
\maketitle
\begin{abstract}
Test-time reinforcement learning enables vision-language models (VLMs) to adapt using unlabeled inputs. However, repeated sampling under fixed visual conditions can reinforce shared perceptual errors, while sequence-level rewards fail to isolate visual perception—the foundational bottleneck that anchors multimodal reasoning—risking the degradation of pre-trained reasoning capabilities. We propose TTRSD, a test-time reinforcement learning framework combining multi-view answer-level self-distillation with visual contrastive token selection. A shared policy aggregates teacher predictions across original, cropped, and downsampled views into an answer distribution. Student trajectories generated from the original image receive rewards based on the support for their final answers in this distribution. To allocate this feedback precisely toward perceptual bottlenecks, we compare the log-probabilities of the same sampled tokens under original and visually ablated inputs while holding their textual prefixes fixed, selecting visually sensitive positions for policy-gradient updates. TTRSD separates update direction, determined by group-relative advantages, from update position, determined by visual sensitivity, without requiring ground-truth labels, external verifiers, or a separate teacher. With only 20 unlabeled adaptation samples, TTRSD improves performance across seven benchmarks and three VLMs, raising InternVL3-2B's MMMU accuracy from 35.79\% to 49.32\% (+13.53\%), demonstrating cross-dataset generalization while preserving inherent reasoning integrity.

\end{abstract}
\begin{figure*}[!ht]
    \centering
    \includegraphics[width=0.85\textwidth]{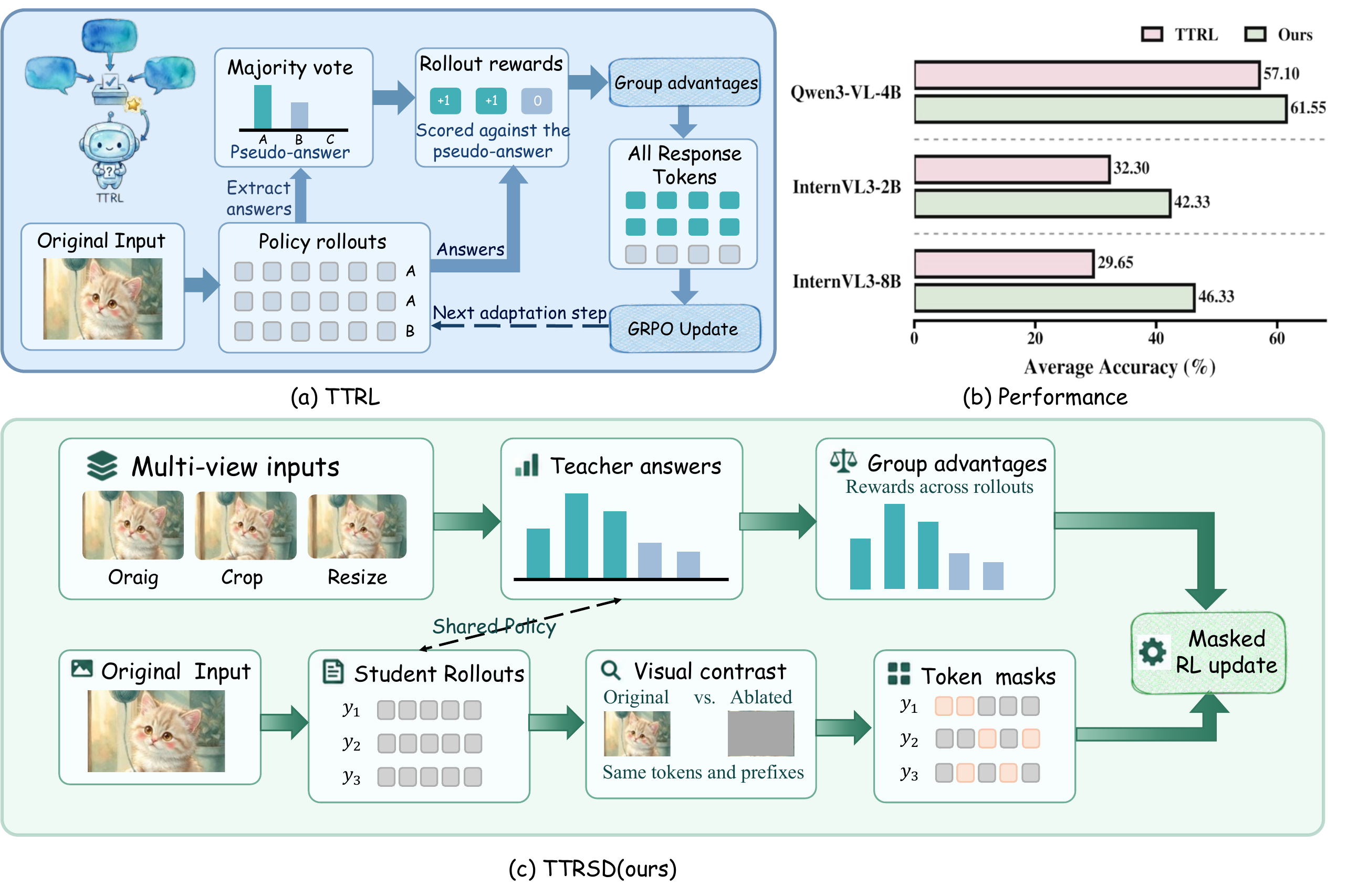} 
    \caption{Overview and performance of \method.
(a) TTRL derives rewards from majority-vote pseudo-labels
and applies sequence-level advantages to all response tokens.
(b) \method~improves average accuracy over TTRL across
seven benchmarks on three VLM backbones.
(c) \method~uses a shared policy to construct rewards
from multi-view teacher answers and selects visually
sensitive tokens through original--ablated prediction
comparisons, enabling targeted policy-gradient updates.}
    \label{fig:intro}
\end{figure*}

\section{Introduction}
Vision-language models (VLMs) have made substantial progress in visual recognition, question answering, and multi-step reasoning  \citep{zhu2025internvl3,bai2025qwen3,li2024llava}. Adapting these models to target tasks in deployment settings without labeled data, however, remains challenging. Test-time reinforcement learning (TTRL) offers a promising approach: it repeatedly samples responses to unlabeled inputs, converts agreement among the model’s own answers into rewards, and updates the policy accordingly. TTRL explores this paradigm for language reasoning \citep{zuo2026ttrl}, TTRV extends it to VLMs \citep{singh2026ttrv}, and TTPO further investigates test-time policy optimization with pseudo-label supervision \citep{wang2026ttpo}. These studies demonstrate the potential of learning from a model’s own predictions, while highlighting the construction and use of supervision as central challenges in unlabeled adaptation.

Representative approaches construct pseudo-supervision
through majority voting or answer frequencies, as
illustrated in Figure~\ref{fig:intro}(a).
Their effectiveness depends on both the quality of
this supervision and how it guides policy updates. Yet repeated sampling under the same visual condition may reproduce shared perceptual errors, allowing incorrect answers to receive strong support and become further reinforced. A central challenge is that answer agreement does not necessarily indicate correct visual understanding, a problem compounded by the fact that additional sampling provides limited diversity in the underlying visual evidence. On-policy self-distillation (OPSD), which derives supervision from a shared model operating under different information conditions, inspires us to broaden supervision sources \citep{zhao2026self,kaur2026rethinking,yu2026preference}. However, privileged teacher information, such as ground-truth answers, is generally unavailable during unlabeled test-time adaptation. Motivated by this limitation, we construct an online teacher using different visual presentations of the same input. By aggregating predictions across views with different spatial coverage and resolutions, we obtain an answer-level self-distillation signal for reasoning over the original image.

Beyond the source of supervision, allocating answer-level feedback across reasoning tokens presents a second challenge. Multimodal reasoning interleaves visual evidence extraction and textual deductions, whose tokens differ in their dependence on visual input. Crucially, errors in visual perception can lead to reasoning errors, while accurate visual grounding supports subsequent reasoning and correct answers. Yet, GRPO broadcasts the same trajectory-level advantage to all
response positions without accounting for these differences.
In test-time adaptation, this feedback is estimated from the model's own predictions on limited unlabeled data. Since final-answer correctness does not necessarily reflect the validity of intermediate visual evidence, applying such feedback throughout the response may penalize valid grounding or reinforce hallucinated evidence.
Furthermore, indiscriminately updating all tokens with sparse and noisy signals exacerbates the stability-plasticity dilemma, risking catastrophic forgetting of the model's robust, pre-trained textual reasoning capabilities. 
Prior work highlights the complementary roles of perception and reasoning~\citep{li2026rethinking} and explore fine-grained optimization using visual dependency~\citep{huang2026spotlight} or token-level teacher predictions~\citep{agarwal2024policy,zhao2026self}, but these methods typically rely on verifiable rewards or available on-policy teachers.
These insights motivate a selective allocation of self-generated feedback during unlabeled adaptation. Inspired by RLSD's separation of update direction and token-level magnitude~\citep{yang2026self}, we use group-relative advantages to determine update directions
and visual sensitivity to select update positions. We identify these positions by comparing token log-probabilities under original and visually ablated inputs with identical textual prefixes, concentrating
policy-gradient updates on predictions more strongly influenced by the image.

As illustrated in Figure~\ref{fig:intro}(c), we propose \method, a test-time reinforcement learning framework combining multi-view answer-level self-distillation with visual contrastive token selection. First, a shared policy serves as both teacher and student. We aggregate teacher predictions from the original, cropped, and downsampled images into an empirical answer distribution. Student trajectories generated from the original image then receive rewards based on voting support for their final answers. Second, inspired by the conditional comparison in visual contrastive decoding~\citep{leng2024mitigating}, we compare the log-probabilities of the same sampled student tokens under original and visually ablated inputs while holding textual prefixes fixed. The absolute differences serve as a visual sensitivity score to select positions for policy-gradient updates. Ultimately, group-relative advantages determine the update direction, while visual sensitivity determines the targeted positions. Reference-policy regularization over all response tokens further constrains overall distribution drift. The entire procedure requires no ground-truth answers, external verifiers, or separate teacher models.

With only 20 unlabeled adaptation samples, \method~ improves performance across seven benchmarks and three VLMs, consistently outperforming the TTRL baseline (\ref{fig:intro}(b)). Maximum gains over zero-shot baselines reach 7.13\% on MMMU for Qwen3-VL-4B, 14.14\% on WeMath for InternVL3-2B, and 15.44\% on MathVista for InternVL3-8B. Notably, the adapted 8B model reaches 71.84\% on MathVista, exceeding the reported results of GPT-4o (71.60\%) and Gemini-2.0-Flash (70.46\%)~\citep{achiam2023gpt,team2023gemini}. Furthermore, positive transfer across evaluated source--target pairs demonstrates strong cross-dataset generalization, while comprehensive ablations confirm the independent effectiveness of both multi-view supervision and visual token selection.

Our contributions are:
\begin{itemize}
    \item We propose \method, a test-time reinforcement
    learning framework that uses multi-view answer-level
    self-distillation to construct rewards without labels
    or a separate teacher.

    \item We introduce visual contrastive token selection
    to decouple update direction from update position:
    teacher answer support determines group-relative
    advantages, while visual sensitivity selects tokens
    for policy-gradient updates.

   \item \method~improves performance across diverse VLMs
and benchmarks with limited unlabeled data and
demonstrates cross-dataset generalization.
\end{itemize}

\section{Related Work}
\noindent
\textbf{Vision--language models (VLMs).}
Vision--language models broadly include dual-encoder models trained with contrastive learning for cross-modal retrieval and recognition~\citep{radford2021learning,jia2021scaling}, and large multimodal models (LMMs) that couple vision encoders with language models for open-ended perception and reasoning~\citep{alayrac2022flamingo,wang2026see,wu2026see,liu2026beyond,du2026medhorizon}. Recent open LMMs, such as Qwen3-VL~\citep{bai2025qwen3} and InternVL3~\citep{zhu2025internvl3}, have substantially improved OCR, spatial perception, long-context 
understanding, and multi-step visual reasoning. Nevertheless, their performance remains sensitive to visual distribution shifts, image quality, and task formats at deployment, while frozen models cannot directly exploit unlabeled test inputs to correct such errors. 
This motivates test-time training (TTT)~\citep{sun2020test}, particularly test-time reinforcement learning (TTRL)~\citep{zuo2026ttrl}, which adapts pretrained models using self-generated supervision from test samples. Following this direction, our method constructs supervision from the same model's 
predictions across multiple visual views, without requiring test-set labels, external verifiers, or an independently trained teacher model.

\noindent
\textbf{Post-training for VLMs.}
Post-training improves language and multimodal models through instruction tuning and reward- or preference-based optimization, including RLHF~\citep{ouyang2022training}, DPO~\citep{rafailov2023direct}, and GRPO~\citep{shao2024deepseekmath}. Multimodal RL methods refine credit assignment at the token level. VPPO~\citep{huang2026spotlight} reshapes trajectory advantages and filters gradients according to visual dependency, while PGPO~\citep{ye2026not} and GCPO~\citep{zhang2026group} calibrate token updates using visual grounding, evidence relevance, or token criticality. A complementary direction studies on-policy distillation: OPD matches teacher predictions on student-generated prefixes~\citep{agarwal2024policy}, OPSD uses a shared model under asymmetric contexts~\citep{zhao2026self}, and RLSD combines reward directions with token-level discrepancies between teacher and student predictions~\citep{yang2026self}. In contrast, TTRSD uses unlabeled test inputs to construct multi-view answer supervision and direct updates toward visually sensitive tokens.

\noindent
\textbf{Test-time training.}
Recently, test-time training (TTT) has emerged as a paradigm that directly adapts model parameters using unlabeled deployment inputs~\citep{sun2020test}. Early studies mainly focused on unimodal classifiers and addressed distribution shifts through self-supervised surrogate objectives such as entropy minimization~\citep{wang2020tent}. Subsequent work extended test-time adaptation to vision-language models, but primarily considered dual-encoder architectures through prompt tuning, image-encoder adaptation, or embedding-space updates~\citep{shu2022test}. More recently, TTRL, TTRV, and TTPO have used answer consensus across sampled responses to construct online rewards, extending test-time policy optimization to generative models~\citep{zuo2026ttrl,singh2026ttrv,wang2026ttpo}. Building on this progression, we propose a view-asymmetric on-policy distillation method that constructs answer-level pseudo-supervision from transformed teacher views and optimizes original-view student trajectories through group-relative reinforcement learning.

\section{Method}
\begin{figure*}[!ht]
    \centering
    \includegraphics[width=1.0\textwidth]{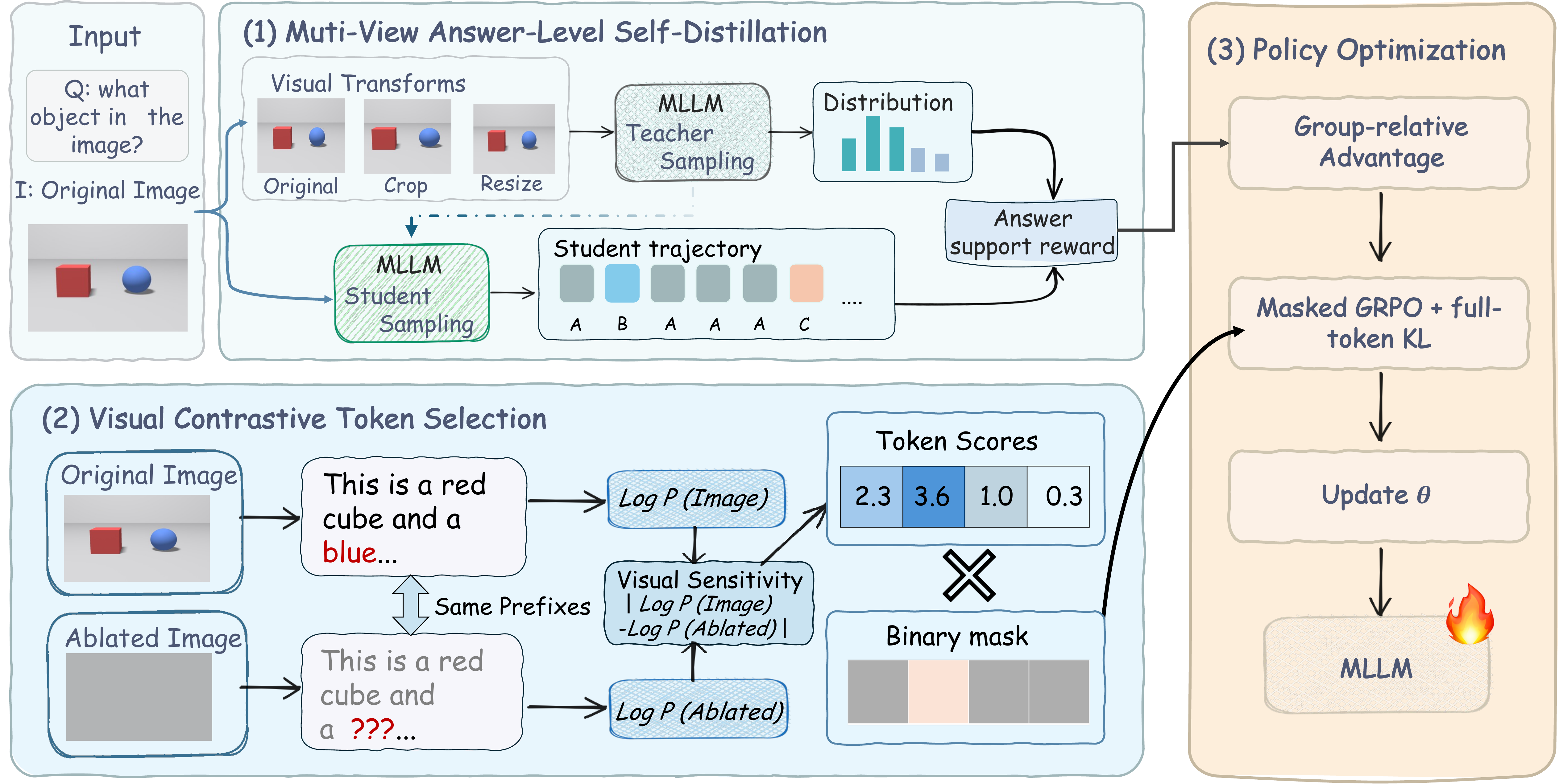} 
    \caption{Overview of \method.
(1) Multi-view answer-level self-distillation aggregates
teacher predictions into an answer distribution to reward
original-view student trajectories.
(2) Visual contrastive token selection compares the
log-probabilities of the same student tokens under original
and ablated images, selecting visually sensitive positions.
(3) Policy optimization combines group-relative advantages
with token masks for GRPO updates, while applying KL
regularization over all valid response tokens.
Teacher and student share the policy parameters
throughout adaptation.}
    \label{fig:method}
\end{figure*}
\subsection{Problem Formulation and Overview}

Let $\pi_\theta$ denote a vision-language model and let
$\mathcal{D}_{\mathrm{adapt}}=\{x_n\}_{n=1}^{N}$ be an unlabeled test-time adaptation set, where each input $x=(q,I)$ consists of a question $q$ and an image $I$. We adapt $\pi_\theta$ using model-generated supervision, without ground-truth answers or a separately trained teacher for optimization.
The shared policy serves as student and teacher under
the following visual contexts:
\begin{equation}
x^S=[q;I], \qquad
x^{T,v}=[q;g_v(I)], \quad v\in\mathcal{V},
\end{equation}
where $g_v$ is a visual transformation, including the identity. The student generates reasoning trajectories
from the original image, while the teacher generates responses across views. As shown in Fig.~\ref{fig:method}, \method~combines two mechanisms.
\textbf{\emph{Multi-view answer-level self-distillation}} aggregates
teacher predictions into an empirical answer distribution
and rewards student trajectories according to the support
for their final answers. \textbf{\emph{Visual contrastive token selection}} compares the likelihoods of the same student tokens under original and image-ablated inputs to identify visually sensitive positions. Together, answer-level supervision determines the group-relative update direction, while visual contrast determines the positions receiving the policy-gradient signal. 

\subsection{Multi-View Answer-Level Self-Distillation}

We construct answer-level supervision from the shared
policy's predictions across visual views and use it to
guide student trajectories generated from the original image.

\noindent
\textbf{Teacher-view construction.}
We consider three visual views,
$\mathcal{V}=\{\mathrm{orig},\mathrm{crop},\mathrm{down}\}$:
\begin{equation}
I^{(\mathrm{orig})}=I,\qquad
I^{(\mathrm{crop})}=\operatorname{Crop}_{0.85}(I),\qquad
I^{(\mathrm{down})}=\operatorname{Resize}_{0.70}(I).
\end{equation}
The center crop retains approximately $85\%$ of each
spatial dimension. Resizing scales both dimensions by
$0.70$, with each resulting dimension lower-bounded by
$64$ pixels. These transformations vary spatial coverage
and resolution to provide alternative visual conditions
for estimating answer support.

\noindent
\textbf{View-asymmetric sampling.}
At adaptation iteration $k$, we sample $N_S$ student
trajectories from the original-view context and $K_T$
teacher trajectories from each teacher-view context:
\begin{equation}
\begin{aligned}
y_i^S &\sim \pi_{\theta_k}(\cdot\mid x^S),
&& i=1,\ldots,N_S,\\
y_j^{T,v} &\sim \pi_{\theta_k}(\cdot\mid x^{T,v}),
&& j=1,\ldots,K_T,\quad v\in\mathcal{V}.
\end{aligned}
\end{equation}

Let $e(\cdot)$ extract the final answer from a completion,
with $a_i^S=e(y_i^S)$ and $a_j^{T,v}=e(y_j^{T,v})$.

\noindent
\textbf{Answer-level supervision.}
The $N_T=|\mathcal{V}|K_T$ teacher answers define an
empirical distribution:
\begin{equation}
\widehat{q}_{T,k}(a\mid x)
=
\frac{1}{|\mathcal{V}|K_T}
\sum_{v\in\mathcal{V}}\sum_{j=1}^{K_T}
\mathbb{I}[a_j^{T,v}=a].
\end{equation}
We reward each student trajectory according to the support for its final answer:
\begin{equation}
r_i^{\mathrm{teach}}
=
\widehat{q}_{T,k}(a_i^S\mid x).
\end{equation}

This reward preserves support for alternative answers
rather than assigning binary rewards based on agreement
with the modal teacher answer. It transfers multi-view
teacher supervision to the original-view student through
policy optimization, without direct token-distribution
matching.

\subsection{Visual Contrastive Token Selection}

Answer-level rewards do not distinguish response tokens by their dependence on visual input. Inspired by visual contrastive decoding~\citep{leng2024mitigating}, we use prediction changes under visual ablation to select visually sensitive positions for policy-gradient updates.

\noindent
\textbf{Visual contrastive scoring.}
For a student trajectory $y=(y_1,\ldots,y_L)$ sampled
from $x^S$, we construct an ablated context
$x^S_{\varnothing}=[q;I_{\varnothing}]$ by zeroing the
processed pixel values while preserving the image-tile
structure and image-token layout. Both forward passes
score the same sampled tokens with identical textual
prefixes; no additional trajectory is generated.
We define token-level visual sensitivity as
\begin{equation}
\Delta_t =
\left|
\log \pi_\theta(y_t\mid x^S,y_{<t})
-
\log \pi_\theta(y_t\mid x^S_{\varnothing},y_{<t})
\right|.
\end{equation}
$\Delta_t$ measures sensitivity to visual
ablation in either direction, rather than token correctness.
The ablated forward pass uses no gradients, and both
log-probabilities are detached for mask construction.
The original-image log-probabilities retain gradients
for policy optimization.

\noindent
\textbf{Token-selective updates.}
Let $L$ denote the number of valid response tokens and
$\tau_\rho$ the $\lceil\rho L\rceil$-th largest score.
The selection mask is
\begin{equation}
M_{\rho,t}
=
\mathbb{I}[\Delta_t\geq\tau_\rho],
\qquad t=1,\ldots,L.
\end{equation}
We apply this rule independently to each student
trajectory with $\rho=0.2$, retaining ties at the threshold.
The mask selects positions for the policy-gradient
objective, while the group-relative advantage determines
the update direction. Reference-policy regularization
remains active over all valid response tokens.

\subsection{Policy Optimization}

Following TTRV~\citep{singh2026ttrv}, we combine teacher support with a diversity-control reward:
\begin{equation}
r_i =
r_i^{\mathrm{teach}}
-\lambda_H\bar H_{T,k}(x),
\end{equation}
where $\bar H_{T,k}(x)$ is the normalized entropy of the empirical teacher answer distribution. And We set $\lambda_H=0.75$ in all experiments.
We first compute group-relative advantages across the
$N_S$ student trajectories for each input, then apply
visual token selection to obtain token-level advantages:
\begin{equation}
A_i =
\frac{r_i-\operatorname{mean}_j(r_j)}
{\operatorname{std}_j(r_j)+\epsilon},
\qquad j=1,\ldots,N_S.
\end{equation}
We then apply the visual mask:
\begin{equation}
\widetilde A_{i,t}=M_{\rho,i,t}A_i.
\end{equation}

Selected tokens retain the sequence-level advantage,
while unselected tokens receive zero policy-gradient
signal. Thus, answer support determines the relative
update direction, and visual sensitivity determines
the update positions.
We incorporate $\widetilde A_{i,t}$ into the GRPO
surrogate~\citep{shao2024deepseekmath}:
\begin{equation}
\mathcal L(\theta)=
-\frac{
\sum_{i,t}\ell_{\mathrm{clip}}
\bigl(u_{i,t}(\theta),\widetilde A_{i,t}\bigr)
}{
\sum_{i,t}M_{\rho,i,t}
}
+\beta\mathcal L_{\mathrm{KL}}(\theta),
\end{equation}
where $u_{i,t}$ is the current-to-old policy probability ratio and $\ell_{\mathrm{clip}}$ is the clipped surrogate, implemented with dual clipping. Sums are taken over valid response positions in each optimization microbatch. $\mathcal L_{\mathrm{KL}}$ regularizes all valid response tokens to constrain overall distribution drift, independently of the visual mask, and  we set $\beta=0.001$. No auxiliary distillation loss is introduced.

\section{Experiments}
\subsection{Setup}
\noindent
\noindent
\textbf{Implementation.}
We adapt InternVL3-2B, InternVL3-8B~\citep{zhu2025internvl3}, and Qwen3-VL-4B~\citep{bai2025qwen3} on 20 unlabeled examples for eight epochs using GRPO, with a learning rate of $5\times10^{-7}$, a clipping parameter of $0.2$, and a KL regularization coefficient of $10^{-3}$. 
All methods, including TTRL and TTRV, use identical rollout and training configurations to ensure a fair comparison. Each rollout batch undergoes one epoch of policy optimization. Ground-truth answers are used only for evaluation and analysis.

\noindent
\textbf{Baselines.}
We consider three categories of baselines.
For closed-source models, we include  GPT-4o ~\citep{achiam2023gpt}and Gemini-2.0-Flash~\citep{team2023gemini}.
For open-source models, we report MiniCPM-V~2.6~\citep{yao2024minicpm}, InternLM-XComposer2-VL-7B~\citep{dong2024internlm}, Phi-3.5-Vision~\citep{haider2024phi}, and LLaVA-1.5-7B~\citep{an2025llava}. For test-time training, we compare against TTRL~\citep{zuo2026ttrl},
which constructs pseudo-rewards through majority voting, and                         TTRV~\citep{singh2026ttrv}, which employs answer-frequency rewards with diversity regularization. The Zero-shot baseline, TTRL, and our method all use chain-of-thought prompting, while TTRV is evaluated both with and without it. All adaptation methods use the same adaptation samples,
training epochs, and total rollout budget.

\noindent
\textbf{Evaluation.}
We evaluate on held-out sets from seven benchmarks:
WeMath~\citep{qiao2025we}, LogicVista~\citep{xiao2024logicvista},
MathVista~\citep{lu2024mathvista}, MathVerse~\citep{zhang2024mathverse},
MathVision~\citep{wang2024measuring}, MMMU~\citep{yue2024mmmu},
and MME-R~\citep{yuan2025mme}.
Zero-shot and post-adaptation evaluations use identical prompts and greedy decoding for each model.
We report accuracy (\%), its unweighted mean across
benchmarks, and gains over zero-shot in percentage
points.

\subsection{Main results} 
Table~\ref{tab:cross-backbone-main} presents the results of \method~
on seven vision--language benchmarks covering visual
mathematics, logical reasoning, and general multimodal
understanding. \method~consistently improves over zero-shot
performance across all benchmarks and backbones, increasing
the average accuracy of Qwen3-VL-4B, InternVL3-2B, and
InternVL3-8B by 4.14\%, 9.96\%, and 7.32\%,
respectively. For example, InternVL3-2B improves by 14.14\%
on WeMath and 13.95\% on LogicVista, while
InternVL3-8B improves from 56.40\% to 71.84\% on MathVista.
Compared with TTRL and TTRV, \method~achieves the highest
average accuracy at all three model scales, demonstrating
consistent gains across tasks and model sizes.

\method~also enables smaller models to compete with larger
counterparts. With only 2B parameters, adapted InternVL3-2B
achieves an average accuracy of 42.33\%, surpassing both
zero-shot InternVL3-8B (39.01\%) and its best TTRV result
(40.65\%). Meanwhile, the 4B Qwen3-VL-4B outperforms the
listed open-source comparison models wherever results
are available across the seven benchmarks, and exceeds
the reported closed-source results on WeMath, MathVista,
MathVerse, and MME-R. Its average accuracy reaches
61.55\%, compared with GPT-4o's 54.46\%. These results
highlight the potential of test-time adaptation to improve
visual reasoning and make smaller models competitive
across diverse tasks. 

\begin{table*}[t]
\caption{Main results on seven vision--language benchmarks.
We report accuracy (\%) and the seven-benchmark average (Avg.).
The best result within each adaptation backbone is shown in
\textbf{bold}. Missing published results are denoted by ``--''.}
\label{tab:cross-backbone-main}
\centering
\resizebox{\textwidth}{!}{%
\begin{tabular}{llcccccccc}
\toprule
\textbf{Backbone} & \textbf{Method}
& \textbf{WeMath} & \textbf{LogicVista}
& \textbf{MathVista} & \textbf{MathVerse}
& \textbf{MathVision} & \textbf{MMMU}
& \textbf{MME-R} & \textbf{Avg.} \\
\midrule
\rowcolor{gray!15}
\multicolumn{10}{c}{\textit{Closed-source Model}} \\
\midrule
GPT-4o
& --
& 50.60 & 64.40 & 71.60
& 49.90 & 43.80 & 70.70
& 30.20 & 54.46 \\
Gemini-2.0-Flash
& --
& 47.42 & -- & 70.46
& 43.65 & 47.82 & 69.30 & -- & --\\
\midrule
\rowcolor{gray!15}
\multicolumn{10}{c}{\textit{Open-source Model}} \\
\midrule
MiniCPM-V 2.6 & -- & 38.62 & 27.54 & 60.60 & 38.30
& 23.40 & 49.80 & 19.34 & -- \\

\shortstack[l]{InternLM-\\XComposer2-VL-7B}
& --
& 12.67 & -- & 57.60 & 25.90
& 14.54 & 43.00 & -- & -- \\

Phi-3.5-Vision
& --
& 11.20 & 25.10 & 43.90 & 36.02
& 14.90 & 43.09 & -- & -- \\

LLaVA-1.5-7B
& --
& -- & 29.24 & 25.40 & 26.02
& 8.52 & 31.30 & 12.10 &  \\

\midrule
\rowcolor{gray!15}
\multicolumn{10}{c}{\textit{Test-time Training Method}} \\
\midrule
\multirow{6}{*}{Qwen3-VL-4B}
& TTRL (w/ CoT)
& 58.28 & 51.30 & 81.06 & 66.51 & \textbf{43.39}
& 58.63 & 40.51 & 57.10 \\
& TTRV (w/o CoT)
& 56.32 & 41.64 & 71.69 & 48.92 & 27.91
& 52.57 & 37.42 & 48.07 \\
& TTRV (w/ CoT)
& 60.06 & 51.06 & 80.00 & 63.25 & 42.86
& 59.25 & 40.82 & 56.76 \\                                                                
& Zero-shot (w/ CoT)
& 71.30 & 56.43 & 79.96 & 60.74 & 35.95
& 53.32 & 44.15 & 57.41 \\
& \method (w/ CoT)
& \textbf{73.58} & \textbf{57.75} & \textbf{81.68}
& \textbf{67.17} & 40.01 & \textbf{60.45}
& \textbf{50.22} & \textbf{61.55} \\
& $\Delta$
& +2.28 & +1.32 & +1.72 & +6.43 & +4.06
& +7.13 & +6.07 & +4.14 \\

\midrule

\multirow{6}{*}{InternVL3-2B}
& TTRL (w/ CoT)
& 12.76 & 24.11 & 57.45 & 49.10 & 19.58
& 47.45 & 15.66 & 32.30 \\
& TTRV (w/o CoT)
& 33.13 & 31.74 & 62.19 & 48.51 & 12.42
& 42.98 & 19.87 & 35.83 \\
& TTRV (w/ CoT)
& 27.64 & 23.48 & 61.43 & 47.44 & 23.76
& 48.44 & 25.31 & 36.79 \\
& Zero-shot (w/ CoT)
& 23.07 & 25.30 & 58.47 & 43.34 & 19.22
& 35.79 & 21.41 & 32.37 \\          
& \method (w/ CoT)
& \textbf{37.21} & \textbf{39.25} & \textbf{66.94}
& \textbf{49.63} & \textbf{27.87} & \textbf{49.32}
& \textbf{26.07} & \textbf{42.33} \\
& $\Delta$
& +14.14 & +13.95 & +8.47 & +6.29 & +8.65
& +13.53 & +4.66 & +9.96 \\

\midrule

\multirow{6}{*}{InternVL3-8B}
& TTRL (w/ CoT)
& 35.69 & 8.51 & 59.57 & 35.78 & 16.60
& 36.02 & 15.35 & 29.65 \\
& TTRV (w/o CoT)
& 43.39 & 20.54 & 56.73 & 21.31 & 26.11
& 46.31 & 32.26 & 35.24 \\
& TTRV (w/ CoT)
& 49.19 & 27.09 & 64.29 & 47.09 & 27.35
& 39.50 & 30.06 & 40.65 \\

& Zero-shot (w/ CoT)
& 35.83 & 32.42 & 56.40 & 50.05 & 25.42
& 44.05 & 28.92 & 39.01 \\

& \method (w/ CoT)
& \textbf{49.54} & \textbf{37.20} & \textbf{71.84}
& \textbf{52.82} & \textbf{31.90} & \textbf{47.02}
& \textbf{34.00} & \textbf{46.33} \\

& $\Delta$
& +13.71 & +4.78 & +15.44 & +2.77 & +6.48
& +2.97 & +5.08 & +7.32 \\

\bottomrule
\end{tabular}%
}
\vspace{2pt}
\end{table*}

\subsection{Ablation Study}

\begin{wrapfigure}[19]{r}{0.50\columnwidth}
    \centering
    \includegraphics[width=\linewidth]{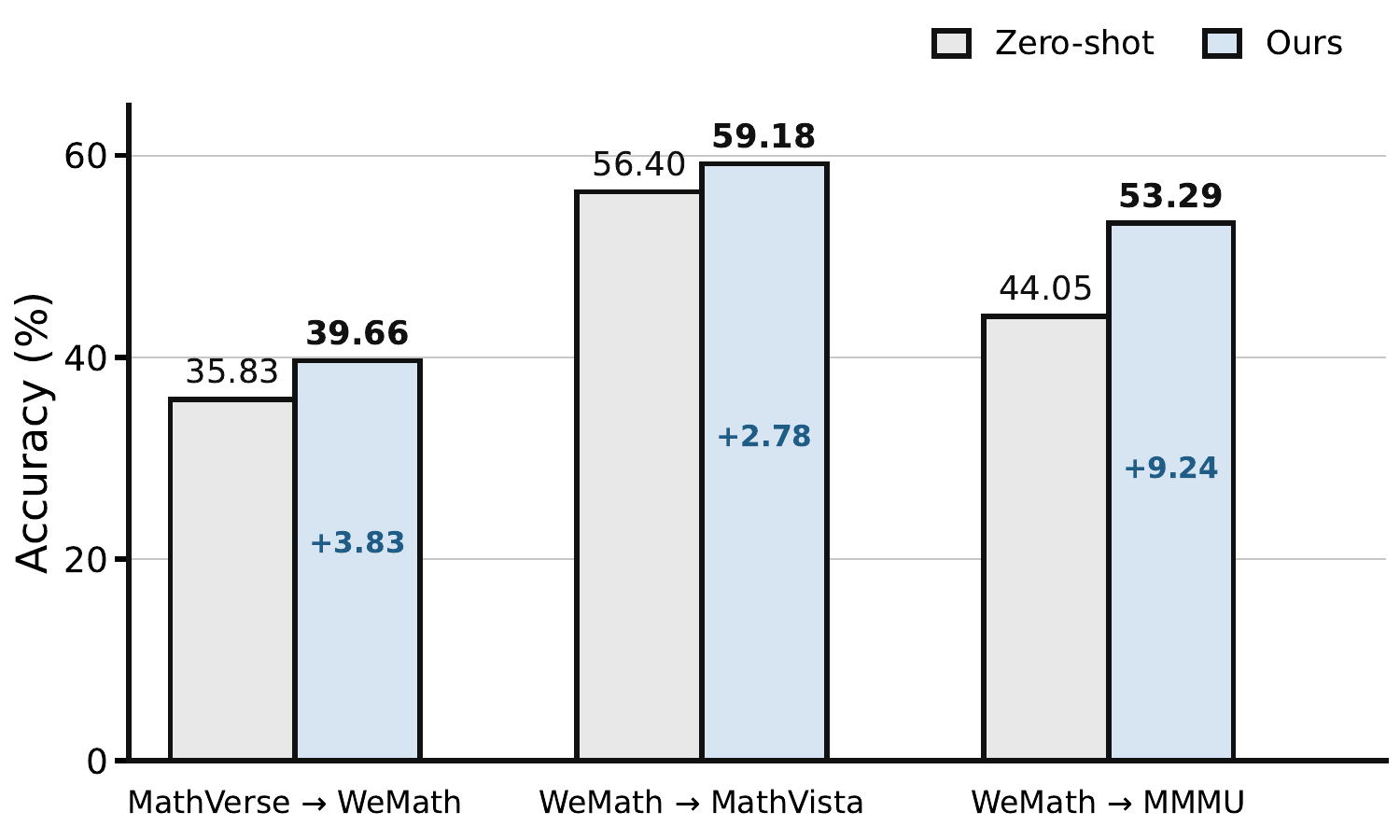}
    \caption{Cross-dataset generalization of \method~with InternVL3-8B.
Bars compare zero-shot and source-adapted accuracy
without target adaptation data.
Arrows indicate transfer direction; gains are in
percentage points.}
    \label{fig:cross-dataset}
\end{wrapfigure}

\textbf{Cross-Dataset Generalization.}
We further evaluate whether 
the benefits of \method~
transfer across datasets. 
For each source-target pair,
we adapt InternVL3-8B exclusively on source data,
without using target inputs or labels during adaptation.
We compare the adapted and zero-shot models on the
same target test set using identical prompts and
greedy decoding.
As shown in Fig.~\ref{fig:cross-dataset}, \method~
improves accuracy across all three transfer settings,
with gains of 3.83 percentage points on
MathVerse$\rightarrow$WeMath and 2.78 points on
WeMath$\rightarrow$MathVista.
The largest gain occurs on WeMath$\rightarrow$MMMU,
where accuracy increases from 44.05\% to 53.29\%,
a gain of 9.24 percentage points.
These results show that the adaptation benefits
extend beyond the source dataset, supporting
the cross-dataset generalization of \method.

\paragraph{Ablation Study on Components.}
We evaluate the contributions of multi-view answer-level
self-distillation (MVSD) and visual contrastive token
selection (VCTS) on MathVista, MathVision, and LogicVista
using InternVL3-2B.
Removing MVSD replaces multi-view teacher sampling with
original-view sampling; removing VCTS applies the
policy-gradient objective to all valid response tokens.
We also evaluate the variant without either component,
keeping the remaining training and evaluation settings fixed.
As shown in Table~\ref{tab:ablation}(a),
the full method achieves the highest accuracy on all
three benchmarks. Removing MVSD reduces accuracy by
4.50, 2.68, and 1.33 percentage points, respectively,
while removing VCTS leads to drops of 5.16, 0.85,
and 8.10 points. Removing both components further
degrades performance, with the largest drop of
15.77 points on LogicVista. These results support
the effectiveness of both components and the
benefit of combining them.

\paragraph{Effect of the token selection ratio.}
We vary the token selection ratio $\rho$ on MathVista
using InternVL3-2B. As shown in Table~\ref{tab:ablation}(b),
accuracy increases from 63.02\% at $\rho=0.1$ to
66.94\% at $\rho=0.2$, then decreases to 63.43\%
and 58.68\% at $\rho=0.3$ and $0.5$, respectively.
Thus, $\rho=0.2$ achieves the best performance among
the evaluated ratios and is used as our default setting.

\begin{table}[t]
\centering
\caption{Ablation studies on InternVL3-2B.
(a) Component ablation: MVSD denotes multi-view
answer-level self-distillation, and VCTS denotes
visual contrastive token selection.
(b) Effect of the token selection ratio $\rho$ on MathVista.
All results are accuracy (\%).}
\label{tab:ablation}
\small
\begin{minipage}[t]{0.55\linewidth}
\centering
(a) Component ablation\par
\smallskip
\setlength{\tabcolsep}{3pt}
\begin{tabular}{ccccc}
\toprule
MVSD & VCTS & MathVista & MathVision & LogicVista \\
\midrule
$\checkmark$ & $\checkmark$
& \textbf{66.94} & \textbf{27.87} & \textbf{39.25} \\
-- & $\checkmark$ & 62.44 & 25.19 & 37.92 \\
$\checkmark$ & -- & 61.78 & 27.02 & 31.15 \\
-- & -- & 61.43 & 23.76 & 23.48 \\
\bottomrule
\end{tabular}
\end{minipage}
\hfill
\begin{minipage}[t]{0.42\linewidth}
\centering
(b) Token selection ratio\par
\smallskip
\setlength{\tabcolsep}{2.5pt}
\begin{tabular}{lccccc}
\toprule
$\rho$ & 0.0 & 0.1 & 0.2 & 0.3 & 0.5 \\
\midrule
Acc. & 58.47 & 63.02 & \textbf{66.94} & 63.43 & 58.68 \\
\bottomrule
\end{tabular}
\end{minipage}
\end{table}

\section{Analysis}
\begin{wrapfigure}[18]{r}{0.5\columnwidth}
    \centering
    \includegraphics[width=\linewidth]
    {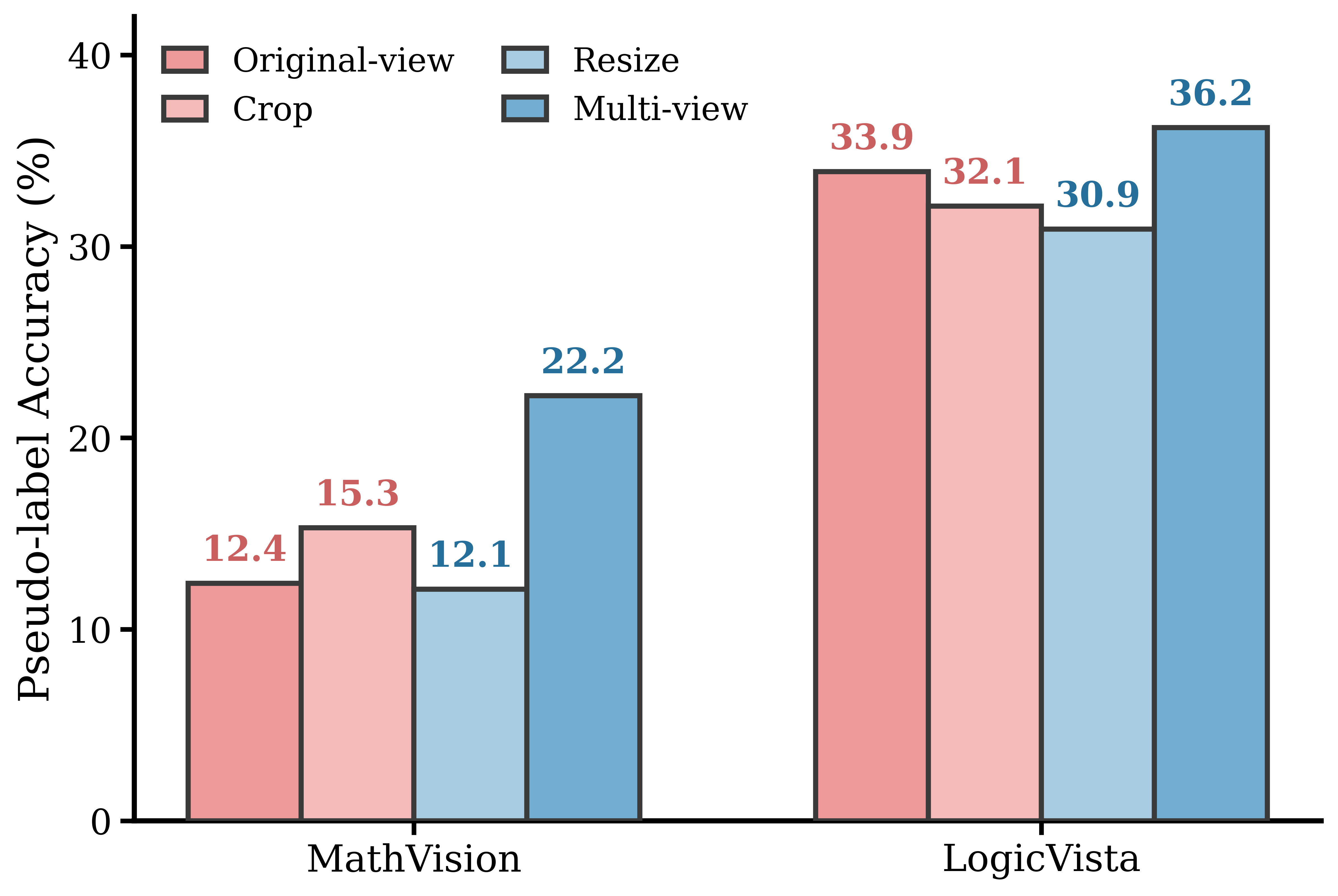}
    \caption{Multi-view teacher voting improves pseudo-label
accuracy over individual views on both benchmarks.}
    \label{fig:multiview_vote_quality}
\end{wrapfigure}

\paragraph{Why transformed teacher views help.}
Figure~\ref{fig:multiview_vote_quality} compares pseudo-label
accuracy from individual teacher views and multi-view
voting. 
On MathVision, multi-view voting achieves 22.2\%,
outperforming the original (12.4\%), cropped (15.3\%),
and resized (12.1\%) views. On LogicVista, it reaches
36.2\%, compared with 33.9\%, 32.1\%, and 30.9\%,
respectively.
Individual transformations do not consistently improve
accuracy: cropping helps on MathVision, while both
transformations underperform the original view on
LogicVista. Nevertheless, aggregating their predictions
outperforms every individual view on both benchmarks.
This suggests that transformed views can provide
complementary predictions even when they are less
accurate in isolation, supporting multi-view aggregation
for more accurate answer-level supervision.

\paragraph{Why visual credit assignment helps.}
Answer-level rewards reflect final prediction correctness but fail to differentiate tokens by their reliance on visual evidence. We evaluate whether visual sensitivity offers a principled criterion for targeted credit assignment. First, we verify the discriminative power of our selection rule. As shown in Table~\ref{tab:visual_credit_assignment}, at $\rho=0.2$, the selected tokens exhibit mean absolute log-probability differences $9.23\times$--$19.11\times$ higher than unselected tokens across six benchmarks. This substantial separation confirms that our contrastive scoring effectively isolates steps directly grounded in visual context from general textual reasoning.

Second, we examine whether this criterion improves adaptation beyond merely reducing the update budget. Table~\ref{tab:token_selection_ablation} compares visual selection against random, high-entropy, and all-token baselines on MathVista under an identical 20\% token budget. Visual selection achieves 66.94\% accuracy, outperforming random and high-entropy selection by 4.49 pp and 5.51 pp, respectively, and surpassing all-token optimization by 5.16 pp. 

\begin{wrapfigure}[22]{r}{0.65\columnwidth}
    \centering
    \includegraphics[width=\linewidth]
    {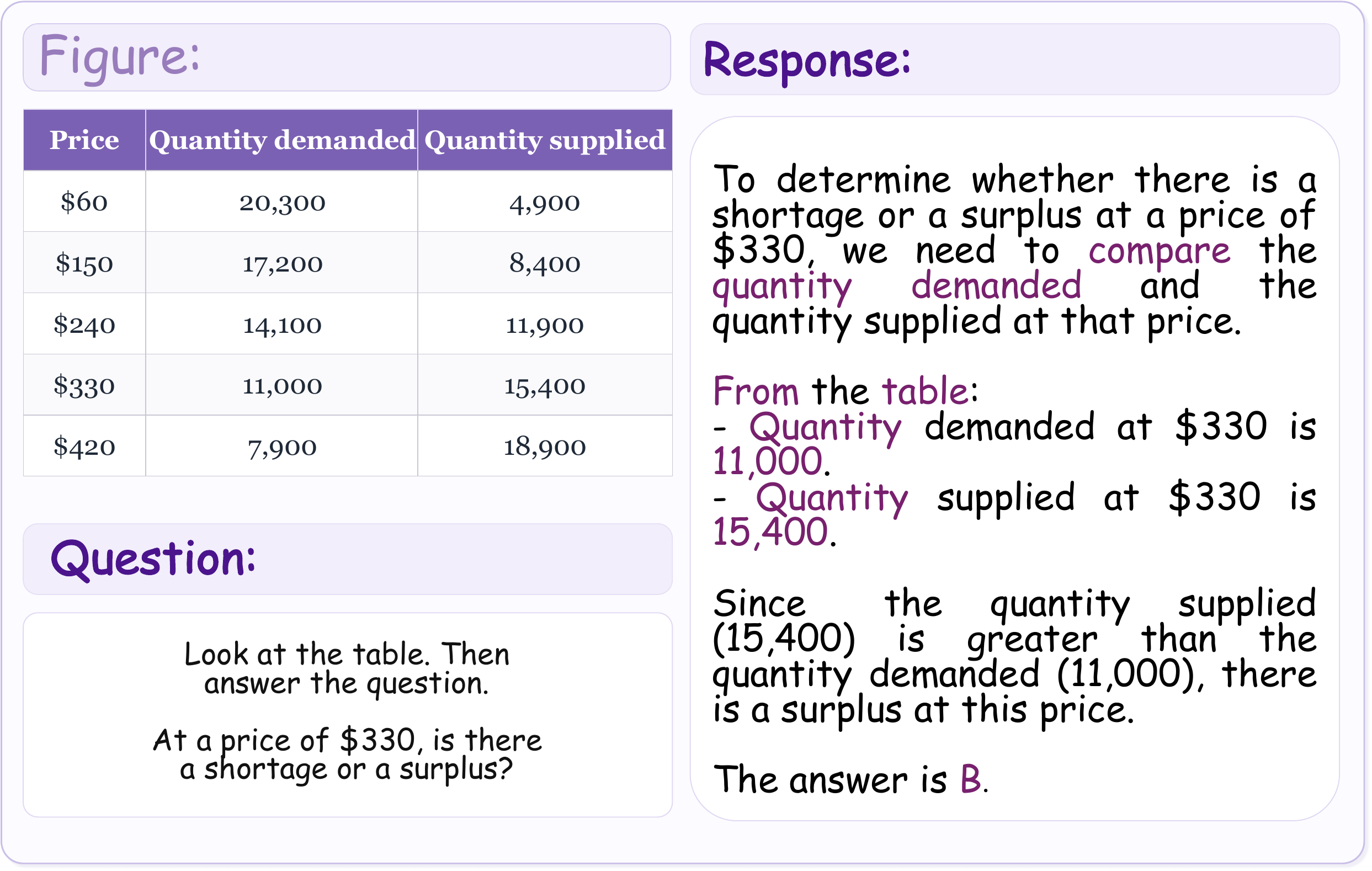}
    \caption{Visual token selection in a table-reading example.
Purple tokens are selected for policy-gradient updates
based on their sensitivity to visual ablation. More visualizations in Appendix.}
    \label{fig:case_study}
\end{wrapfigure}
These comparisons highlight the importance of the
token-selection criterion.
\textbf{High-entropy selection} underperforms all-token
optimization by 0.35 pp, indicating that prioritizing
uncertain tokens does not improve adaptation in this
setting.
\textbf{Random selection} yields a modest gain of
0.67 pp, showing that reducing the number of updated
positions can provide some benefit even without
visual guidance.
Visual selection achieves a larger gain of 5.16 pp
and outperforms both alternatives under the same
20\% token budget.
These results support visual dependence as an
effective criterion for selecting which response
positions receive answer-level policy-gradient feedback.

\begin{table}[t]
\centering
\begin{minipage}[t]{0.52\linewidth}
    \vspace{0pt}
    \centering
    \caption{Token-level visual sensitivity at $\rho=0.2$.
    Values are mean absolute log-probability differences
    between original and visually ablated inputs.}
    \label{tab:visual_credit_assignment}
    \small
    \setlength{\tabcolsep}{4pt}
    \begin{tabular}{lrrr}
        \toprule
        Benchmark & All & Selected & Unselected \\
        \midrule
        WeMath     & 0.1709 & 0.7012 & 0.0367 \\
        LogicVista & 0.4283 & 1.5862 & 0.1262 \\
        MathVista  & 0.3825 & 1.4089 & 0.0974 \\
        MathVerse  & 0.3538 & 1.1875 & 0.1286 \\
        MathVision & 0.2036 & 0.7295 & 0.0677 \\
        MMMU       & 0.2839 & 1.0563 & 0.0837 \\
        \bottomrule
    \end{tabular}
\end{minipage}
\hfill
\begin{minipage}[t]{0.45\linewidth}
    \vspace{0pt}
    \centering
    \caption{Token-selection ablation on MathVista.
    Selection methods retain 20\% of valid response tokens;
    other training settings are fixed.}
    \label{tab:token_selection_ablation}
    \small
    \setlength{\tabcolsep}{4pt}
\begin{tabular}{lrr}
    \toprule
    Selection rule & Acc.\ (\%) & $\Delta$ (pp) \\
    \midrule
    Visual (ours) & \textbf{66.94} & \textbf{+5.16} \\
    Random        & 62.45 & +0.67 \\
    High entropy  & 61.43 & -0.35 \\
    All tokens    & 61.78 & 0.00 \\
    \bottomrule
\end{tabular}
\end{minipage}
\end{table}

\paragraph{Case study.}
Figure~\ref{fig:case_study} visualizes token selection
in a table-reading example, with selected tokens shown
in purple.
The selection captures tokens in the initial extraction
of ``11,000'' and ``15,400'', together with descriptive
tokens such as ``quantity'' and ``table''.
Later repetitions of these values remain unselected,
consistent with their availability in the textual prefix.
This example illustrates how visual sensitivity
distinguishes the same content at different positions
in a reasoning trajectory.

\section{Conclusion}
We presented \method, a test-time reinforcement learning
framework that enables VLMs to improve from their own
predictions without ground-truth labels, external
verifiers, or a separate teacher.
By combining multi-view answer-level self-distillation
with visual contrastive token selection, \method~
uses teacher answer support to determine update
direction and visual sensitivity to select update
positions. 
Experiments across seven benchmarks and three VLMs
demonstrate consistent performance gains and effective
cross-dataset generalization. 
This work suggests a path toward VLM self-evolution
through multi-view self-supervision and selective
policy updates.

\bibliography{conference}
\bibliographystyle{conference}

\clearpage

\appendix
\section{Implementation Details}
We provide the complete training configuration in Table~\ref{tab:training_hyperparameters}. All experiments employ on-policy training with bfloat16 precision, FSDP, gradient checkpointing, and padding removal. We train for eight epochs using an actor learning rate of 
$5\times10^{-7}$, a PPO clipping range of $0.2$, one PPO epoch per batch, and a KL-loss coefficient of $10^{-3}$. During training, 
responses are sampled with a temperature of $1.0$ and top-$p$ of $1.0$, whereas evaluation uses greedy decoding. We set the maximum 
prompt and completion lengths to 7,524 and 3,072 tokens, respectively. The hard visual mask retains the top $20\%$ most visually 
sensitive completion tokens for policy-gradient updates, while KL regularization remains active over all valid completion tokens. No 
ground-truth reward, confidence filtering, or self-paced sample selection is used in the principal experiments.
    \begin{table}[!ht]
    \centering
    \caption{Training hyperparameters used in all experiments.}
    \label{tab:training_hyperparameters}
    \small
    \begin{tabular}{lc}
    \toprule
    \textbf{Hyperparameter} & \textbf{Value} \\
    \midrule
    Actor learning rate & $5\times10^{-7}$ \\
    Training epochs & 8 \\
    PPO epochs per batch & 1 \\
    PPO clipping range & 0.2 \\
    KL-loss coefficient & $1\times10^{-3}$ \\
    Training batch size & 32 \\
    Rollouts per input & 16 \\
    Sampling temperature & 1.0 \\
    Top-$p$ & 1.0 \\
    Maximum prompt length & 7,524 \\
    Maximum response length & 3,072 \\
    Numerical precision & bfloat16 \\
    Visual-token ratio $\rho$ & 0.2 \\
    \bottomrule
    \end{tabular}
    \end{table}

\section{Multi-Seed Experiments}
\label{app:multi_seed}
To evaluate the robustness of our method to stochastic variation, we conduct multi-seed experiments on five benchmarks using 
Qwen3-VL-4B. For each benchmark, we repeat the complete adaptation procedure with five independent random seeds. The adaptation 
examples, evaluation splits, prompts, and training hyperparameters are fixed across runs, while the random seed controls stochastic rollout generation, data ordering, and policy optimization. Table~\ref{tab:multi_seed_results} reports the mean accuracy and standard 
deviation over the five runs.

The results demonstrate that TTRSD produces consistent performance across different random seeds. The standard deviation does 
not exceed $1.26$ percentage points on any benchmark, and remains below one percentage point on four of the five benchmarks. Overall,  the limited variation across independent runs indicates that the reported performance is robust to training stochasticity and is not driven by a particular favorable random seed.

\begin{table}[!ht]
\centering
\caption{Multi-seed results of TTRSD with Qwen3-VL-4B. Accuracy (\%) is reported as the mean and standard deviation over five
independent random seeds.}
\label{tab:multi_seed_results}
\small
\begin{tabular}{lc}
\toprule
\textbf{Benchmark} & \textbf{Accuracy (\%)} \\
\midrule
LogicVista & $ 57.58\pm 0.17$ \\
MathVerse  & $67.01 \pm 0.16$ \\
MathVision & $ 39.64\pm 0.37$ \\
MathVista  & $ 80.93\pm 0.75$ \\
WeMath     & $ 72.96\pm 1.26$ \\
\bottomrule
\end{tabular}
\end{table}

\section{Scaling to Larger Adaptation Sets}
\label{app:data_scaling}
To further examine the scalability of our method, we investigate how its performance changes as the amount of unlabeled adaptation data increases. Specifically, we compare adaptation using 20 and 100 questions on WeMath, MMMU, and LogicVista with InternVL3-2B. All
experiments follow the same multi-view construction, chain-of-thought prompting, pseudo-label generation, and visual token optimization
procedures.

As shown in Table~\ref{tab:data_scaling}, increasing the adaptation set consistently improves performance across all three benchmarks. Accuracy increases from 37.21\% to 38.14\% on WeMath, from 49.32\% to 50.21\% on MMMU, and from 39.25\% to 41.78\% on LogicVista, corresponding to gains of 0.93, 0.89, and 2.53 percentage points, respectively. These results indicate that our method can effectively exploit additional unlabeled examples rather than relying on a small collection of favorable adaptation instances. A larger adaptation set exposes the model to a broader range of visual content and reasoning patterns,
thereby providing more diverse self-supervision for policy optimization.

The performance improvement is accompanied by additional computational cost, since a larger adaptation set requires more multi-view rollout
generation, pseudo-label estimation, and policy updates. Therefore, the number of adaptation examples introduces a practical trade-off
between adaptation performance and computational efficiency. Overall, the consistent improvements observed across all three benchmarks demonstrate that multi-view pseudo-labeling and visual token selection remain effective as the amount of unlabeled adaptation data increases.

\begin{table}[!ht]
\centering
\caption{Effect of scaling the unlabeled adaptation set from 20 to 100 questions with InternVL3-2B. Accuracy is reported in percent.}
\label{tab:data_scaling}
\small
\begin{tabular}{lccc}
\toprule
\textbf{Benchmark}
& \textbf{20 Questions}
& \textbf{100 Questions}
& $\boldsymbol{\Delta}$ \\
\midrule
WeMath     & 37.21 & 38.14 & 0.93 \\
MMMU       & 49.32 & 50.21 & 0.89 \\
LogicVista & 39.25 & 41.78 & 2.53 \\
\bottomrule
\end{tabular}
\end{table}

\section{Results on Perception-Oriented and Multi-Disciplinary Benchmarks}
\label{app:perception_benchmarks}

Beyond mathematical reasoning, we further evaluate our method on
perception-oriented and multi-disciplinary benchmarks, including texture classification (DTD),  multi-disciplinary perception (MMStar and SEED-Bench), and real-world scene understanding (RealWorldQA). As reported in Table~\ref{tab:perception_benchmarks}, our method consistently improves over the zero-shot baseline on all of these benchmarks, with a particularly large gain of $+50.8$ points on DTD. These results indicate that the benefit of on-policy adaptation with our method is not confined to mathematical reasoning, but generalizes across diverse multimodal task categories.

\begin{table}[!ht]
\centering
\caption{Results on perception-oriented and multi-disciplinary benchmarks
with InternVL3-2B. Accuracy (\%) is reported before and after adaptation.}
\label{tab:perception_benchmarks}
\small
\begin{tabular}{lccc}
\toprule
\textbf{Benchmark} & \textbf{Zero-shot} & \textbf{Ours} & \textbf{$\Delta$} \\
\midrule
DTD         & 37.12 & \textbf{87.94} & +50.8 \\
MMStar      & 47.97 & \textbf{51.11} & +3.1  \\
SEED-Bench  & 69.85 & \textbf{70.82} & +1.0  \\
RealWorldQA & 63.75 & \textbf{64.57} & +0.8  \\
\bottomrule
\end{tabular}
\end{table}

\section{Computational Overhead Analysis}
\label{app:overhead}
The only component our method adds over the TTRV baseline is one additional teacher-forced forward pass under the blank-image condition; no extra autoregressive rollout is required. We profile the wall-clock cost of each stage of a training step on a single NVIDIA A100 GPU with InternVL3-2B in bfloat16, using a representative sequence (a 919-token prompt covering three image tiles and a 111-token response), and report the average over ten runs in Table~\ref{tab:overhead}. The blank-image forward pass costs 94.2\,ms,
essentially identical to its real-image counterpart (94.1\,ms), since the two differ only in the visual input. In practice, autoregressive rollout generation dominates the training step, accounting for over 96\% of the end-to-end time, whereas the blank-image pass contributes less than 1\%.
The overhead introduced by visual sensitivity estimation is therefore negligible.
 
\begin{table}[t]
\centering
\caption{Wall-clock breakdown of one training step for a single sequence
(InternVL3-2B, bfloat16, one NVIDIA A100 GPU; averaged over ten runs). The
blank-image scoring pass is the only component added by our method.}
\label{tab:overhead}
\small
\begin{tabular}{lrr}
\toprule
\textbf{Stage} & \textbf{Time} & \textbf{Share} \\
\midrule
Rollout generation (111 tokens, autoregressive) & 13{,}170\,ms & 96.5\% \\
Real-image scoring forward (no grad)            & 94.1\,ms     & 0.7\%  \\
Blank-image scoring forward (no grad, ours)     & 94.2\,ms     & 0.7\%  \\
Policy forward + backward                       & 224.4\,ms    & 1.6\%  \\
Optimizer update                                & 63.4\,ms     & 0.5\%  \\
\midrule
Total                                           & 13{,}646\,ms & 100\%  \\
\bottomrule
\end{tabular}
\end{table}

\section{Limitations and Future Work}

\textbf{Dependence on multi-view voting.}
Our method relies on multi-view responses to construct pseudo-labels. When all views produce correlated errors or no rollout 
yields the correct answer, the resulting supervision may be unreliable. Dynamically weighting views according to their agreement 
could mitigate this limitation.

\noindent
\textbf{Fixed visual token selection.}
We retain a fixed proportion of visually sensitive tokens based on the difference between real-image and blank-image likelihoods.
 However, the optimal proportion may vary across questions. Future work could adapt the selection ratio according to visual 
dependence and question difficulty.

\noindent
\textbf{Domain scope and efficiency.}
Our experiments focus on vision--language reasoning tasks with extractable answers. Extending the method to open-ended tasks may 
require semantic answer clustering or learned verification. Moreover, multi-view sampling and counterfactual forward passes 
introduce additional computation. Dynamically allocating views and rollouts according to pseudo-label confidence could improve 
adaptation efficiency.

\section{Prompt Template}
\label{app:prompt_template}
We first present the general prompt template used for rollout generation and evaluation. The image, question, and answer choices provided by each benchmark are inserted into the corresponding placeholders. We then provide one representative prompt from each benchmark to illustrate the specific question and answer-choice formats.

\subsection{General Prompt Template}

\begin{tcolorbox}[
title={Chain-of-Thought Prompt}
]
\ttfamily
\textless Image\textgreater

\textless Question\textgreater

\textless Answer Choices\textgreater

Reason concisely using the visual evidence and the choices.
Use enough steps to avoid guessing, but keep the reasoning focused.
End with a separate final line exactly in the form:
The answer is X.
\end{tcolorbox}

\subsection{Benchmark-Specific Examples}

\begin{tcolorbox}[
enhanced,
breakable,
width=\linewidth,
colback=promptbackground,
colframe=promptgreen,
boxrule=0.8pt,
arc=2mm,
left=4mm,
right=4mm,
top=5mm,
bottom=3mm,
title={WeMath Prompt},
fonttitle=\bfseries,
coltitle=white,
attach boxed title to top left={
xshift=4mm,
yshift=-2mm
},
boxed title style={
colback=promptdarkgreen,
colframe=promptdarkgreen,
boxrule=0pt,
arc=1.5mm,
left=3mm,
right=3mm,
top=1.5mm,
bottom=1.5mm
}
]
\ttfamily
\textless Image\textgreater

\medskip
\textless Question\textgreater

As shown in the diagram, there is a circular dining table with an
approximate diameter of 8 dm. To cover this dining table with a square
tablecloth, the store offers three sizes of tablecloths. The side
length of the tablecloth should be at least (~~~~).

\medskip
\textless Answer Choices\textgreater

A. 15 decimeters

B. 7 decimeters

C. 9 decimeters

D. No correct answer

\medskip
Reason concisely using the visual evidence and the choices.
Use enough steps to avoid guessing, but keep the reasoning focused.
End with a separate final line exactly in the form:
The answer is X.
\end{tcolorbox}

\begin{tcolorbox}[
enhanced,
breakable,
width=\linewidth,
colback=promptbackground,
colframe=promptgreen,
boxrule=0.8pt,
arc=2mm,
left=4mm,
right=4mm,
top=5mm,
bottom=3mm,
title={LogicVista Prompt},
fonttitle=\bfseries,
coltitle=white,
attach boxed title to top left={
xshift=4mm,
yshift=-2mm
},
boxed title style={
colback=promptdarkgreen,
colframe=promptdarkgreen,
boxrule=0pt,
arc=1.5mm,
left=3mm,
right=3mm,
top=1.5mm,
bottom=1.5mm
}
]
\ttfamily
\textless Image\textgreater

\medskip
\textless Question\textgreater

The scale is balanced. What weight is the weight with the question
mark? Select from A, B, C, D, and E.

\medskip
\textless Answer Choices\textgreater

A. 1 lb

B. 2 lb

C. 8 lb

D. 10 lb

E. 25 lb

\medskip
Reason concisely using the visual evidence and the choices.
Use enough steps to avoid guessing, but keep the reasoning focused.
End with a separate final line exactly in the form:
The answer is X.
\end{tcolorbox}

\begin{tcolorbox}[
enhanced,
breakable,
width=\linewidth,
colback=promptbackground,
colframe=promptgreen,
boxrule=0.8pt,
arc=2mm,
left=4mm,
right=4mm,
top=5mm,
bottom=3mm,
title={MathVista Prompt},
fonttitle=\bfseries,
coltitle=white,
attach boxed title to top left={
xshift=4mm,
yshift=-2mm
},
boxed title style={
colback=promptdarkgreen,
colframe=promptdarkgreen,
boxrule=0pt,
arc=1.5mm,
left=3mm,
right=3mm,
top=1.5mm,
bottom=1.5mm
}
]
\ttfamily
\textless Image\textgreater

\medskip
\textless Question\textgreater

In the figure above, triangle ABC is inscribed in the circle with
center O and diameter AC. If AB = AO, what is the degree measure of
angle ABO?

\medskip
\textless Answer Choices\textgreater

A. 15 degrees

B. 30 degrees

C. 45 degrees

D. 60 degrees

E. 90 degrees

\medskip
Reason concisely using the visual evidence and the choices.
Use enough steps to avoid guessing, but keep the reasoning focused.
End with a separate final line exactly in the form:
The answer is X.
\end{tcolorbox}

\begin{tcolorbox}[
enhanced,
breakable,
width=\linewidth,
colback=promptbackground,
colframe=promptgreen,
boxrule=0.8pt,
arc=2mm,
left=4mm,
right=4mm,
top=5mm,
bottom=3mm,
title={MathVerse Prompt},
fonttitle=\bfseries,
coltitle=white,
attach boxed title to top left={
xshift=4mm,
yshift=-2mm
},  
boxed title style={
colback=promptdarkgreen,
colframe=promptdarkgreen,
boxrule=0pt,
arc=1.5mm,
left=3mm,
right=3mm,
top=1.5mm,
bottom=1.5mm
}
]
\ttfamily
\textless Image\textgreater

\medskip
\textless Question\textgreater      

Emile is observing a wind turbine. The vertical distance between the
ground and the tip of one of the turbine's blades, in meters, is
modeled by \(H(t)\), where \(t\) is the time in seconds. What is the
meaning of the highlighted segment?

\medskip
\textless Answer Choices\textgreater

A. The turbine's center is 35 meters above the ground.

B. The turbine completes a single cycle in 35 seconds.

C. The length of the blade is 35 meters.

D. The turbine has 35 blades.

\medskip
Reason concisely using the visual evidence and the choices.
Use enough steps to avoid guessing, but keep the reasoning focused.
End with a separate final line exactly in the form:
The answer is X.
\end{tcolorbox}

\begin{tcolorbox}[
enhanced,
breakable,
width=\linewidth,
colback=promptbackground,
colframe=promptgreen,
boxrule=0.8pt,
arc=2mm,
left=4mm,
right=4mm,
top=5mm,
bottom=3mm,
title={MathVision Prompt},
fonttitle=\bfseries,
coltitle=white,
attach boxed title to top left={
xshift=4mm,
yshift=-2mm
},
boxed title style={
colback=promptdarkgreen,
colframe=promptdarkgreen,
boxrule=0pt,
arc=1.5mm,
left=3mm,
right=3mm,
top=1.5mm,
bottom=1.5mm
}
]
\ttfamily
\textless Image\textgreater

\medskip
\textless Question\textgreater

\(ABCD\) is a square, and \(M\) and \(N\) are the midpoints of
\(BC\) and \(CD\), respectively. Then \(\sin\theta=\)

\medskip
\textless Answer Choices\textgreater

A. \(\frac{\sqrt{5}}{5}\)

B. \(\frac{3}{5}\)

C. \(\frac{\sqrt{10}}{5}\)

D. \(\frac{4}{5}\)

E. None of these

\medskip
Reason concisely using the visual evidence and the choices.
Use enough steps to avoid guessing, but keep the reasoning focused.
End with a separate final line exactly in the form:
The answer is X.
\end{tcolorbox}

\begin{tcolorbox}[
enhanced,
breakable,
width=\linewidth,
colback=promptbackground,
colframe=promptgreen,
boxrule=0.8pt,
arc=2mm,
left=4mm,
right=4mm,
top=5mm,
bottom=3mm,
title={MMMU Prompt},
fonttitle=\bfseries,
coltitle=white,
attach boxed title to top left={
xshift=4mm,
yshift=-2mm
},  
boxed title style={
colback=promptdarkgreen,
colframe=promptdarkgreen,
boxrule=0pt,
arc=1.5mm,
left=3mm,
right=3mm,
top=1.5mm,
bottom=1.5mm
}
]
\ttfamily
\textless Image\textgreater

\medskip
\textless Question\textgreater      

A hash table of length 10 uses open addressing with the hash function
\(h(k)=k\bmod 10\) and linear probing. After inserting six values into
an empty hash table, the table is as shown in the image. Which option
gives a possible order in which the key values could have been
inserted?

\medskip
\textless Answer Choices\textgreater

A. 46, 42, 34, 52, 23, 33

B. 34, 42, 23, 52, 33, 46

C. 46, 34, 42, 23, 52, 33

D. 42, 46, 33, 23, 34, 52

\medskip
Reason concisely using the visual evidence and the choices.
Use enough steps to avoid guessing, but keep the reasoning focused.                                                   
End with a separate final line exactly in the form:
The answer is X.
\end{tcolorbox}

\begin{tcolorbox}[
enhanced,
breakable,
width=\linewidth,
colback=promptbackground,
colframe=promptgreen,
boxrule=0.8pt,
arc=2mm,
left=4mm,
right=4mm,
top=5mm,
bottom=3mm,
title={MME-Reasoning Prompt},
fonttitle=\bfseries,
coltitle=white,
attach boxed title to top left={
xshift=4mm,
yshift=-2mm
},  
boxed title style={
colback=promptdarkgreen,
colframe=promptdarkgreen,
boxrule=0pt,
arc=1.5mm,
left=3mm,
right=3mm,
top=1.5mm,
bottom=1.5mm
}
]
\ttfamily
\textless Image\textgreater

\medskip
\textless Question\textgreater

As shown in the diagram, the flower color of a certain plant is
controlled by two pairs of independently inherited alleles on
autosomes: \(D,d\) and \(R,r\). Which of the following statements is
incorrect?

\medskip
\textless Answer Choices\textgreater

A. There are four genotypes that can produce purple flowers.

B. When plants with genotype \(DdRr\) are self-crossed, the proportion
of purple-flowered offspring that can stably inherit the trait is
\(\frac{1}{6}\).

C. When a \(Ddrr\) plant is crossed with a \(ddRR\) plant, half of the
offspring have blue flowers and half have purple flowers.

D. When a \(DDrr\) plant is crossed with a \(ddRr\) plant, half of the
offspring have blue flowers and half have purple flowers.

\medskip
Reason concisely using the visual evidence and the choices.
Use enough steps to avoid guessing, but keep the reasoning focused.
End with a separate final line exactly in the form:
The answer is X.
\end{tcolorbox}

\section{Additional Case Studies}
  \label{app:case_studies}
As a complement, we provide additional qualitative case studies, including positive cases and a failure case, for a more comprehensive analysis of our method. For each example, the same greedy response is re-scored with the real image and with a blank image, and the tokens in the top-$20\%$ of $\Delta$ are highlighted, visualizing how the selected tokens correspond to the visual evidence in the image. These examples also present the behavior of our method in both successful and failed cases, providing complementary insight for a comprehensive understanding of our method.
\begin{figure*}[!ht]
    \centering
    \includegraphics[width=0.70\textwidth]{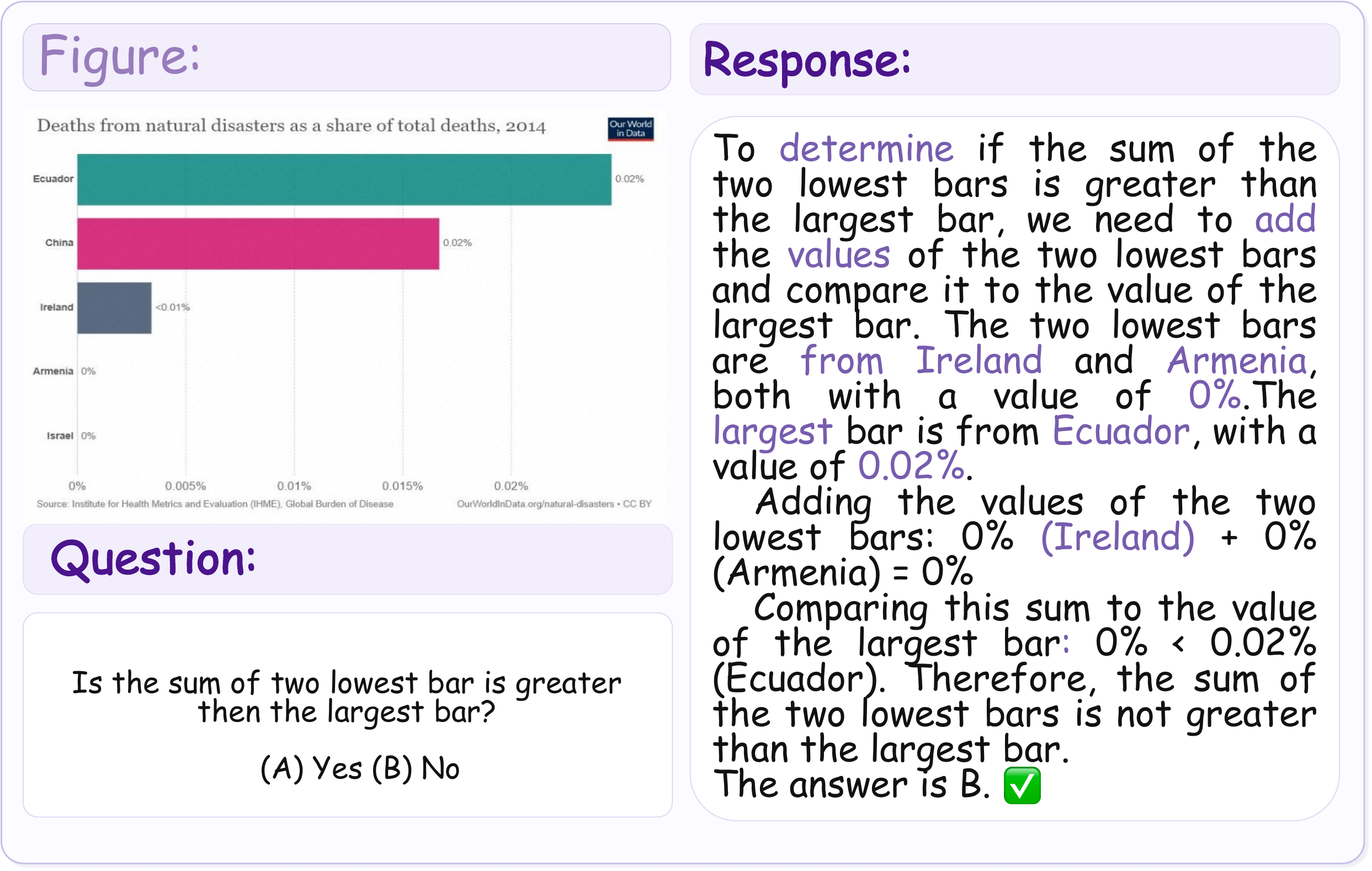} 
    \caption{\textbf{Case Study 1:} numeric extraction from a chart.}
    \label{fig:intro}
\end{figure*}
\begin{figure*}[!ht]
    \centering
    \includegraphics[width=0.70\textwidth]{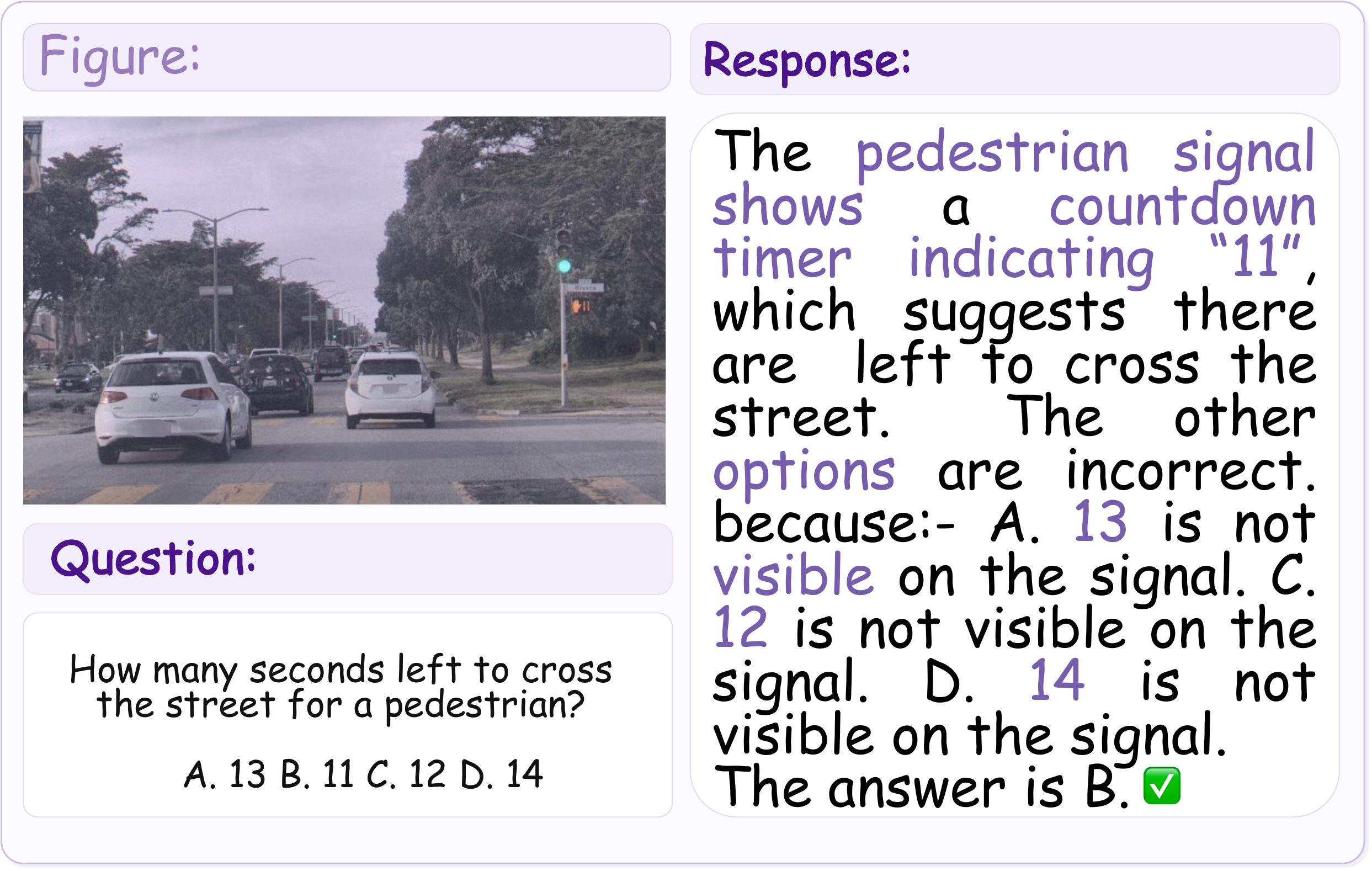} 
    \caption{\textbf{Case Study 2:} object counting. }
    \label{fig:intro}
\end{figure*}
\begin{figure*}[!ht]
    \centering
    \includegraphics[width=0.70\textwidth]{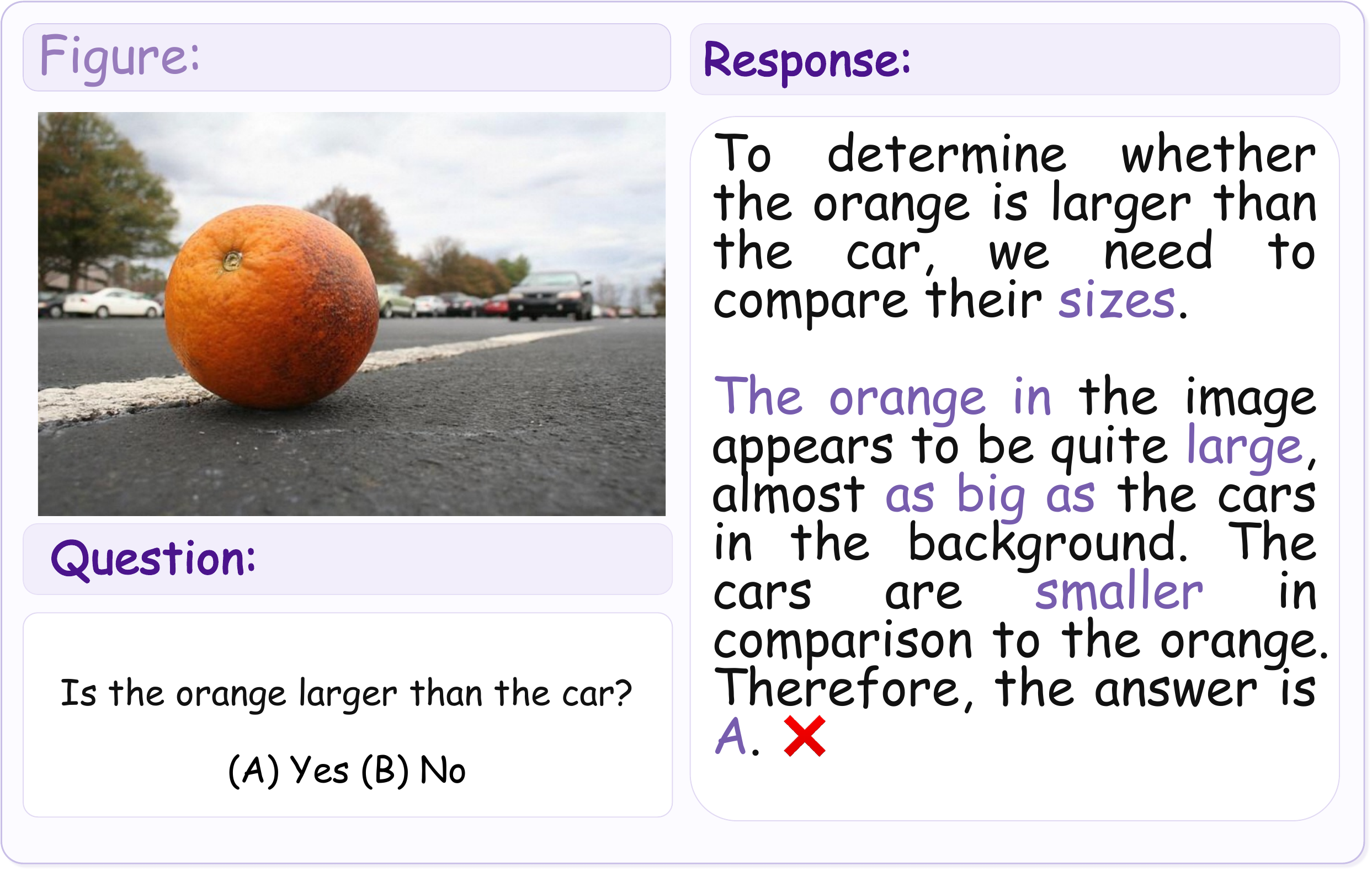} 
    \caption{\textbf{Case Study 3 (failure):} a perceptual illusion.}
    \label{fig:intro}
\end{figure*}

\section{Overall Adaptation Algorithm}
\label{sec:overall_algorithm}

We summarize the complete adaptation procedure of \method~through three components. Algorithm~\ref{alg:teacher_distribution} constructs answer-level pseudo-supervision from multiple teacher views.
Algorithm~\ref{alg:visual_selection} identifies the response tokens whose probabilities are most sensitive to the visual input. Algorithm~\ref{alg:ttopsd} combines these components to optimize
original-view student trajectories using GRPO.

The teacher distribution is estimated exclusively from transformed
teacher-view responses. Student responses therefore do not participate
in constructing their own training targets, reducing the direct
self-confirmation that arises from single-view majority voting.

The resulting mask measures visual dependence rather than token
uncertainty. It is applied only to the policy-gradient component,
whereas KL regularization remains active over the complete valid
response sequence.

In the reported configuration, we use $N_S=16$ student rollouts and
$K_T=16$ samples for each of the three teacher views, yielding
48 teacher votes and 64 total responses per question. We set
$\rho=0.2$, $\lambda_H=0.75$, $\beta=10^{-3}$, and use an actor
learning rate of $5\times10^{-7}$. The model is optimized for eight
epochs with a PPO clipping range of $0.2$ and one PPO epoch per batch.
Only original-view student trajectories receive policy-gradient
updates, while teacher-view trajectories are used exclusively to
construct pseudo-supervision.

\begin{algorithm}[t]
\caption{Multi-View Teacher Distribution}
\label{alg:teacher_distribution}
\begin{algorithmic}[1]
\Require Example $x=(q,I)$, policy $\pi_\theta$,
teacher prompt $p_T$, samples per view $K_T$
\Ensure Teacher answer distribution $q_T(\cdot\mid x)$

\State Construct the teacher-view set
\[
\mathcal{V}(I)
\gets
\left\{
I,\,
\operatorname{Crop}_{0.85}(I),\,
\operatorname{Resize}_{0.70}(I)
\right\}
\]
\State Initialize $\mathcal{Y}_T\gets\varnothing$

\ForAll{$I^{(v)}\in\mathcal{V}(I)$}
    \State Sample teacher rollouts
    \[
    \mathcal{Y}_T^{(v)}
    \gets
    \left\{
    y_{v,j}^{T}
    \sim
    \pi_\theta(\cdot\mid p_T,q,I^{(v)})
    \right\}_{j=1}^{K_T}
    \]
    \State
    $\mathcal{Y}_T
    \gets
    \mathcal{Y}_T\cup\mathcal{Y}_T^{(v)}$
\EndFor

\State Construct the teacher answer distribution
\[
q_T(a\mid x)
\gets
\frac{1}{|\mathcal{Y}_T|}
\sum_{y\in\mathcal{Y}_T}
\mathbb{I}[e(y)=a]
\]

\State \Return $q_T(\cdot\mid x)$
\end{algorithmic}
\end{algorithm}

\begin{algorithm}[t]
\caption{Counterfactual Visual Token Selection}
\label{alg:visual_selection}
\begin{algorithmic}[1]
\Require Question $q$, image $I$, response $y$,
policy $\pi_\theta$, selection ratio $\rho$
\Ensure Visual-token mask $M^\rho$

\State Construct the blank-image counterfactual $I_{\varnothing}$
by zeroing the image pixels while preserving the visual-token layout

\For{$t=1,\ldots,|y|$}
\State Compute visual sensitivity
\[
\Delta_t
\gets
\left|
\log\pi_\theta(y_t\mid q,I,y_{<t})
-
\log\pi_\theta
(y_t\mid q,I_{\varnothing},y_{<t})
\right|
\]
\EndFor

\State Select the largest $\lceil\rho|y|\rceil$ scores
\[
M^\rho
\gets
\operatorname{TopMask}
\left(
\{\Delta_t\}_{t=1}^{|y|},
\lceil\rho|y|\rceil
\right)
\]  

\State \Return $M^\rho$
\end{algorithmic}
\end{algorithm}

\begin{algorithm}[!t]
\caption{\method: Test-Time On-Policy Self-Distillation}
\label{alg:ttopsd}
\begin{algorithmic}[1]
\Require Unlabeled adaptation set $\mathcal{D}_{\mathrm{adapt}}$;
initial policy $\pi_{\theta_0}$; student and teacher prompts
$p_S,p_T$; rollout numbers $N_S,K_T$; coefficients
$\rho,\lambda_H,\beta$
\Ensure Adapted policy $\pi_\theta$

\State Initialize $\theta\gets\theta_0$

\For{each adaptation epoch}
\ForAll{$x=(q,I)\in\mathcal{D}_{\mathrm{adapt}}$}
    \State Set rollout policy
    $\pi_{\mathrm{old}}\gets\pi_\theta$

    \State Construct teacher supervision
    \[
    q_T(\cdot\mid x)
    \gets
    \Call{TeacherDistribution}                                                                                    
    {x,\pi_{\mathrm{old}},p_T,K_T}
    \]

    \State Sample original-view student rollouts
    \[
    \mathcal{Y}_S
    \gets
    \left\{
    y_i^S
    \sim
    \pi_{\mathrm{old}}(\cdot\mid p_S,q,I)
    \right\}_{i=1}^{N_S}
    \]

    \ForAll{$y_i^S\in\mathcal{Y}_S$} 
        \State Assign trajectory reward
        \[
        r_i
        \gets
        q_T(e(y_i^S)\mid x)
        -
        \lambda_H\bar H(q_T)
        \]

        \State Construct visual-token mask
        \[
        M_i^\rho
        \gets
        \Call{VisualTokenSelection}
        {q,I,y_i^S,\pi_\theta,\rho}
        \]
    \EndFor

    \State Compute group-relative advantages                                                                      
    \[
    \{A_i\}_{i=1}^{N_S}
    \gets
    \operatorname{GRPOAdvantage}
    \left(\{r_i\}_{i=1}^{N_S}\right)
    \]

    \State Compute the visually masked objective
    \[
    \mathcal{L}_{\method}
    \gets
    -
    \frac{
    \displaystyle
    \sum_{i,t}
    M_{i,t}^{\rho}
    \ell_{\mathrm{clip}}(\theta;A_i)
    }{
    \displaystyle
    \sum_{i,t}M_{i,t}^{\rho}
    }
    +
    \beta\mathcal{L}_{\mathrm{KL}}
    \]

    \State Update the policy
    \[
    \theta
    \gets
    \theta-\eta\nabla_\theta\mathcal{L}_{\method}
    \]
\EndFor
\EndFor

\State \Return $\pi_\theta$
\end{algorithmic}
\end{algorithm}

\end{document}